\documentclass[sigconf]{aamas} 

\usepackage{balance} 

\usepackage{amsfonts}
\usepackage{xcolor}
\usepackage{amsmath}
\usepackage{subcaption}

\newcommand{\mdp}{\mathcal{M}}
\newcommand{\xt}{\mathbf{X}_t}
\newcommand{\yt}{\mathbf{Y}_t}
\newcommand{\yHatt}{\hat{\mathbf{Y}}_t}

\usepackage[ruled, linesnumbered, noend]{algorithm2e}

\SetCommentSty{mycommfont}
\newcommand{\argmax}{\mathop{\mathrm{argmax}}}

\setcopyright{ifaamas}
\acmConference[AAMAS '21]{Proc.\@ of the 20th International Conference on Autonomous Agents and Multiagent Systems (AAMAS 2021)}{May 3--7, 2021}{Online}{U.~Endriss, A.~Now\'{e}, F.~Dignum, A.~Lomuscio (eds.)}
\copyrightyear{2021}
\acmYear{2021}
\acmDOI{}
\acmPrice{}
\acmISBN{}

\acmSubmissionID{667}

\title{Minimum-Delay Adaptation in Non-Stationary Reinforcement Learning via Online High-Confidence Change-Point Detection}

\author{Lucas N. Alegre}
\affiliation{
  \institution{Institute of Informatics - Federal University of Rio Grande do Sul}
  \city{Porto Alegre, RS - Brazil}}
\email{lnalegre@inf.ufrgs.br}

\author{Ana L. C. Bazzan}
\affiliation{
  \institution{Institute of Informatics - Federal University of Rio Grande do Sul}
  \city{Porto Alegre, RS - Brazil}}
\email{bazzan@inf.ufrgs.br}

\author{Bruno C. da Silva}
\affiliation{
  \institution{CICS - University of Massachusetts}
  \city{Amherst, MA}}
\email{bsilva@cs.umass.edu}

\begin{abstract}
Non-stationary environments are challenging for reinforcement learning algorithms. If the state transition and/or reward functions change based on latent factors, the agent is effectively tasked with optimizing a behavior that maximizes performance over a possibly infinite random sequence of Markov Decision Processes (MDPs), each of which drawn from some unknown distribution. We call each such MDP a \textit{context}. 
Most related works make strong assumptions such as knowledge about the distribution over contexts, the existence of pre-training phases, or \textit{a priori} knowledge about the number, sequence, or boundaries between contexts. 
We introduce an algorithm that efficiently learns policies in non-stationary environments. It analyzes a possibly infinite stream of data and computes, in real-time, high-confidence change-point detection statistics that reflect whether novel, specialized policies need to be created and deployed to tackle novel contexts, or whether previously-optimized ones might be reused. We show that \textit{(i)} this algorithm minimizes the delay until unforeseen changes to a context are detected, thereby allowing for rapid responses; and \textit{(ii)} it bounds the rate of false alarm, which is important in order to minimize regret.
Our method constructs a mixture model composed of a (possibly infinite) ensemble of probabilistic dynamics predictors that model the different modes of the distribution over underlying latent MDPs.
We evaluate our algorithm on high-dimensional continuous reinforcement learning problems and show that it outperforms state-of-the-art (model-free and model-based) RL algorithms, as well as state-of-the-art meta-learning methods specially designed to deal with non-stationarity.
\end{abstract}

\keywords{Reinforcement Learning; Non-stationarity; Model-based RL; Change-Point Detection}

\newcommand{\BibTeX}{\rm B\kern-.05em{\sc i\kern-.025em b}\kern-.08em\TeX}

\begin{document}


\pagestyle{fancy}
\fancyhead{}


\maketitle 


\section{Introduction}

Reinforcement learning (RL) techniques have been successfully applied to solve high-dimensional sequential decision problems. However, if the state transition and/or reward functions change unexpectedly,  according to latent factors unobservable to the agent, the system is effectively tasked with optimizing behavior policies that maximize performance over a (possibly infinite) random sequence of Markov Decision Processes (MDPs). Each MDP is drawn from an unknown distribution and is henceforth referred to as a \textit{context}. This is known as a non-stationary setting. Designing efficient algorithms to tackle this problem is a known challenge in RL \cite{Padakandla2020}. The key difficulties here result from \textit{(i)} the need to quickly and reliably detect when the underlying system dynamics has changed; and \textit{(ii)} the need to effectively learn and deploy adaptable prediction models and policies, specialized in particular contexts, while allowing the agent to (when appropriate) reuse previously-acquired knowledge. Non-stationary environments often result from systems whose dynamics are inherently time-dependent; from agents that are tasked with learning policies under noisy or missing sensors; or from problems where the agent faces sequences of unlabeled/unidentified tasks, each one with its own transition dynamics and reward functions.

Non-stationary settings arise naturally in many situations. Humans, for instance, are capable of learning to solve sequences of tasks from few experiences while preserving knowledge from older experiences \cite{Lake2015}. Consider, for example, a person realizing the need to adapt their gait after an accident, learning novel gait patterns to use when walking with crutches, and then, after a period of recovery, successfully re-deploying normal walking gaits. This corresponds to a non-stationary scenario where the agent needs to learn specialized dynamics models and policies for tackling different contexts/learning scenarios.

We introduce an algorithm that efficiently learns decision-making strategies in this setting. It assumes that an agent experiences random sequences of contexts (MDPs) drawn from some unknown distribution, and it is capable of optimizing behaviors even when a pre-training phase (during which the agent interacts with sample contexts) is not available. The agent's goal is to rapidly deploy policies that are appropriate for each randomly-arriving context---even when the number of latent contexts is unknown and when the context distribution cannot be modeled nor controlled by the agent. We are particularly interested in the case where quick readaptation and sample efficiency are paramount to achieving good performance; for instance, in novel contexts where collecting experiences and acquiring policies from scratch is unfeasible.

Many existing related works tackle non-stationary problems either by detecting when the underlying MDP changes, or via meta-learning approaches that construct a prior model (or policy) capable of rapidly generalizing to novel contexts. Hadoux et al., for example, introduced a technique based on change-point detection algorithms to deal with non-stationary problems with discrete state spaces \cite{Hadoux+2014,Banerjee+2017}. We, by contrast, address the more general setting of high-dimensional continuous RL problems. Supervised meta-learning algorithms \cite{Finn+2017} have also been recently combined with RL to enable fast adaptation under changing domains \cite{Nagabandi+2019a, Nagabandi+2019b}. Nagabandi et al. introduced a model-based algorithm where a meta-learning technique is used to construct dynamics models capable of rapidly adapting to recent experiences---either by updating the hidden state of a recurrent neural network \cite{Duan+2016}, or by updating the model parameters via a small number of gradient steps \cite{Finn+2017}. Meta-learning methods typically assume disjoint training and testing phases, so that an agent can be pre-trained over randomly sampled contexts prior to its deployment. We, however, do not require a pre-training phase. Meta-learning methods also typically assume that the distribution over contexts experienced during training is the same as the one experienced during testing, so that agents can adapt to novel environments with structural similarities to those previously experienced. We, by contrast, do not require that contexts are sampled from a previously-seen distribution, nor that contexts share structural similarities with previously-experienced ones.

To address these limitations, we introduce an algorithm that analyzes a possibly infinite stream of data and computes, in real-time, high-confidence change-point detection statistics that reflect whether novel, specialized policies need to be deployed to tackle new contexts, or whether a previously-optimized policy may be reused. We call our algorithm \underline{M}odel-\underline{B}ased RL \underline{C}ontext \underline{D}etection, or MBCD. We formally show that it minimizes the delay until unforeseen changes to a context are detected, thereby allowing for rapid responses, and that it allows for formal bounds on the rate of false alarm---which is of interest when minimizing the agent's regret over random sequences of contexts. Our method constructs a mixture model composed of a (possibly infinite) ensemble of probabilistic dynamics predictors that model the different modes of the distribution over underlying latent MDPs. 
Our method is capable of optimizing policies based on streams of arbitrarily different contexts, with unknown boundaries, and which may be drawn from an arbitrary, unknown distribution.

We evaluate our algorithm on high-dimensional continuous reinforcement learning tasks and empirically show \textit{(i)} that state-of-the-art reinforcement learning algorithms struggle to deal with non-stationarity; and \textit{(ii)} that our method outperforms state-of-the-art meta-learning methods specifically tailored to deal with non-stationary environments---in particular, when the agent is faced with MDPs that are off-distribution with respect to the set of training contexts provided to the agent, or when novel contexts are structurally different from previously-observed ones. 

\section{Problem Formulation}
\label{sec:problem}

We define a non-stationary environment as a family of MDPs $\{\mdp_{z}\}_{z \in \mathbb{N}^+}$. Each MDP $\mdp_z$ is  a tuple $(\mathcal{S}, \mathcal{A}, \mathcal{T}_z, \mathcal{R}_z, \gamma, d^0)$, where $\mathcal{S}$ is a (possibly continuous) state space, $\mathcal{A}$ is a (possibly continuous) action space, $\mathcal{T}_z: \mathcal{S} \times \mathcal{A} \times \mathcal{S} \rightarrow [0,1]$ is a transition function specifying the distribution over next states, given the current state and action, $\mathcal{R}_z: \mathcal{S} \times \mathcal{A} \rightarrow \mathbb{R}$ is a reward function, $\gamma \in [0,1]$ is the discount factor, and $d^0$ is an initial state distribution. In what follows, $S_t$, $A_t$, and $R_t$ are the random variables corresponding to the state, action, and reward at time step $t$. We assume that the agent observes a random sequence $(\mdp_0,\mdp_1,\ldots)$ of MDPs---called \textit{contexts}---drawn independently from some unknown distribution. We assume that the number of contexts, $|\{\mdp_{z}\}|$, is unknown. These definitions are similar to those discussed in \cite{Banerjee+2017} and \cite{Padakandla2020}.  
Let $z$ be a latent index variable indicating a particular MDP, $\mdp_z$. We assume that each MDP's transition and reward function are parameterized by a latent vector $\theta_z$. Let $p_{\theta_z}(S_{t+1},R_t|S_t,A_t)$ denote the joint conditional probability distribution over next-state and reward associated with MDP $\mdp_z$. We do not impose any smoothness assumptions on how variations to $\theta_z$ affect $\mathcal{T}_z$ and $\mathcal{R}_z$: contexts may be arbitrarily different and share no structural similarities.

Let the time steps in which context changes occur be an increasing sequence of integer random variables, $\{C_i\}_{i\geq1}$, for which a prior $\phi(C_i)$ is unknown or cannot be defined. We call each $C_i$ a \textit{change-point}. At every change-point $C_i$, the current context $\mdp_z$ is replaced by a new randomly drawn MDP. To perform well, an agent must rapidly detect context changes and deploy an appropriate policy. If a new random context differs significantly from previously-experienced ones, the agent may have to learn a policy from scratch; otherwise, it may choose to reuse previously-acquired knowledge to accelerate learning and avoid catastrophic forgetting.

At each time $t$, when interacting with its environment, the agent selects an action $A_t$ based on its state $S_t$  according to a stochastic policy $\pi(\cdot|S_t)$. Let $v^H_{\pi, \mdp_z}$ be the value function associated with policy $\pi$, MDP $\mdp_z$, and defined over a horizon $H$:
\begin{equation}
    \label{eq:return}
    v^H_{\pi, \mdp_z}(s) =  \mathbb{E} \left[\sum_{j=0}^{H-1} \gamma^j R_{t+j} | S_t = s, \pi, \mdp_z \right].
\end{equation}

To simplify our analysis, we first consider (without loss of generality) the simpler case of a family of MPDs $\{\mdp_0, \mdp_1\}$, where a context change occurs from $\mdp_0$ to $\mdp_1$ at some unobserved random time $C$. The mathematical arguments that follow can be extended to the more general setting with an arbitrary number of contexts. 
Let $\Pi^*$ be a policy that follows the optimal policy for $\mdp_0$, $\pi_0$, before $C$, and the optimal policy for $\mdp_1$, $\pi_1$, afterwards. This policy's value function is defined as:
\begin{equation}
\label{eq:optimal}
    v^{\infty}_{\Pi^*,\{\mdp_0,\mdp_1\}}(s) = \mathbb{E} \left[ v_{\pi_0,\mdp_0}^{C}(s) + \gamma^{C} \mathbb{E}_{s'\sim \rho^{\pi_0}_{\mdp_0, C-1}} \left[ v_{\pi_1,\mdp_1}^{\infty}(s')\right] \right],
\end{equation}
\noindent where $\rho^{\pi}_{\mdp, t}$ denotes the distribution of  states reachable after following policy $\pi$ for $t$ steps under MDP $\mdp$. 
Notice that Eq.~\ref{eq:optimal} models the case when $\mdp_1$ starts in the (random) state where $\mdp_0$ terminated, that is, immediately prior to the random time $C$. This implies that $d^0_{\mdp_1} = \rho^{\pi_0}_{\mdp_0, C-1}$.

By contrast, consider an alternative policy $\Pi$ that follows policy $\pi_1$ only after a random time $\Gamma$, for $\Gamma > C$; that is, a policy that deploys the correct decision-making strategy for $\mdp_1$ with a delay of $\Delta = \Gamma - C$ steps. Its value function is given by: 
\begin{multline}
\label{eq:valuedelay}
    v^{\infty}_{\Pi,\{\mdp_0,\mdp_1\}}(s) = \mathbb{E} [v_{\pi_0,\mdp_0}^{C}(s) + \\
    \gamma^{C} \mathbb{E}_{s'\sim \rho^{\pi_0}_{\mdp_0,C-1}} \left[v_{\pi_0, \mdp_1}^{\Delta}(s') \right] + \\
    \gamma^{C+\Delta+1} \mathbb{E}_{s'\sim \rho^{\pi_0}_{\mdp_1, \Delta}} \left[v_{\pi_1,\mdp_1}^{\infty}(s') \right]].
\end{multline}
Notice that we can rewrite Eq.~\ref{eq:optimal} in the same form as Eq.~\ref{eq:valuedelay}:
\begin{multline}
\label{eq:optimal2}
    v^{\infty}_{\Pi^*,\{\mdp_0,\mdp_1\}}(s) = \mathbb{E} [v_{\pi_0,\mdp_0}^{C}(s) + \\
    \gamma^{C} \mathbb{E}_{s'\sim \rho^{\pi_0}_{\mdp_0,C-1}} \left[v_{\pi_1, \mdp_1}^{\Delta}(s') \right] + \\
    \gamma^{C+\Delta+1} \mathbb{E}_{s'\sim \rho^{\pi_1}_{\mdp_1, \Delta}} \left[v_{\pi_1,\mdp_1}^{\infty}(s') \right]].
\end{multline}

We now define the regret $\mathcal{L}(\Delta)$ as the expected discounted sum of rewards lost due to the delay $\Delta = \Gamma - C$, when changing from $\pi_0$ to $\pi_1$ only at time step $\Gamma$. This quantity is given by the difference between Eq.~\ref{eq:valuedelay} and Eq.~\ref{eq:optimal2}. Since we are interested in minimizing the delay $\Delta$, we assume the adversarial case when policies $\pi_0$ and $\pi_1$ do, in the short-term (i.e., within the delay window $\Delta$), have a nearly indistinguishable distribution over the states that are reachable in $\Delta$ steps. In particular, we assume that the KL divergence between $\rho_{\mdp_1,\Delta}^{\pi_0}$ and $\rho_{\mdp_1,\Delta}^{\pi_1}$ is bounded and small for small values of $\Delta$. In this case, the regret can be approximated by\footnote{This definition can be extended in a straightforward way to the case where there is a sequence of random change-points, $\{C_i\}_{i\geq1}$. In particular, the regret will be defined in terms not of a single random delay, but of a sequence of random delays associated with the corresponding random contexts that are observed by the agent.}:
\begin{equation}
\label{eq:regret}
    \mathcal{L}(\Delta) \approx \mathbb{E} \left[ \gamma^{C} \mathbb{E}_{s'\sim \rho^{\pi_0}_{\mdp_0, \Gamma-1}} \left[v_{\pi_1, \mdp_1}^{\Delta}(s') - v_{\pi_0, \mdp_1}^{\Delta}(s')\right] \right]
\end{equation}
\noindent as $\mathcal{D}_{KL}(\rho_{\mdp_1,\Delta}^{\pi_0}||\rho_{\mdp_1,\Delta}^{\pi_1}) \to 0$. From Eq.~\ref{eq:regret}, it should be clear that to maximize the expected return over the random sequence of MDPs, one needs to minimize regret; and to minimize regret, one needs to minimize the random delay $\Delta$. In the next section, we introduce a method capable of minimizing the worst-case delay until unforeseen changes to a stochastic process are detected, while also bounding the rate of false alarm---i.e., the likelihood that the method will \textit{incorrectly} indicate that a context change occurred.

\section{High-Confidence Change-Point Detection}
\label{sec:cusum}

Change-point detection (CPD) algorithms \cite{AminikhanghahiCook2016,Veeravalli+2012} are designed to detect whether (and when) a change occurs in the distribution generating random observations from an arbitrary stochastic process. These methods have been widely used in a variety of fields---from financial markets \cite{Lam+1997} to biomedical signal processing \cite{Sibanda+2007}.
Although CPDs have been applied to reinforcement learning problems \cite{Banerjee+2017, Hadoux+2014}, the application and formal analysis of such methods have been restricted to discrete state spaces settings.

In the \textit{online} CPD setting, a sequential detection procedure is defined to rapidly and reliably estimate when the parameter $\theta$ of some underlying distribution or stochastic process has changed. Online CPD algorithms should produce high-confidence estimates, $\Gamma$, of the true change-point time, $C$. Notice that $\Gamma$ is a random variable whose stochasticity results from the unknown stochastic prior over context changes, $\phi$, and from the fact that each MDP in $\{\mdp_{z}\}$ produces random trajectories of states, actions, and rewards.

Suppose that at each time $t$, while the agent interacts with $\mdp_0$, sample next-state and rewards are drawn from $p_{\theta_0}(S_{t+1}, R_t|S_t, A_t)$, where $\theta_0$ is the latent vector parameterizing $\mdp_0$'s transition and reward functions. At some unknown random change-point $C$, the context changes to $\mdp_1$, and experiences that follow are drawn from $p_{\theta_1}(S_{t+1}, R_t|S_t, A_t)$. We propose to identify such a change by computing high-confidence statistics that reflect whether $\theta_0$ has changed.
This can be achieved by introducing a minimax formulation of the CPD problem, as discussed by Pollak \cite{Pollak1985}. In this formulation, the goal of a CPD algorithm is to compute a random estimate, $\Gamma$, of the latent change-point time $C$, such that \textit{(i)} it minimizes the worst-case expected detection delay, $\Delta_{worst}(\Gamma)$, associated with the random estimates $\Gamma$ produced by a particular CPD algorithm, when considering all possible change-points $C$; and \textit{(ii)} bounds on the maximum false alarm rate (FAR) may be imposed. The worst-case expected detection delay, $\Delta_{worst}(\Gamma)$, and the FAR, are defined as:
\begin{equation}
\label{eq:pollak}
    \Delta_{worst}(\Gamma) = \sup_{c \geq 1} \ \mathbb{E} [\Gamma - C | \Gamma \geq C, C=c],
\end{equation}

\begin{equation}
\label{eq:far}
    \operatorname{FAR}(\Gamma) = \frac{1}{\mathbb{E}[\Gamma|C=\infty]},
\end{equation}
\noindent where the expectations in Eq.~\ref{eq:pollak} and Eq.~\ref{eq:far} are over the possible histories of experiences produced by the stochastic process, and where conditioning on $C = \infty$ indicates the random event where the context never changes.
Given these definitions, the objective of a high-confidence change-point detection process is the following:
\begin{equation}
    \underset{\Gamma}{\inf} \ \Delta_{worst}(\Gamma) \ \text{subject to} \ \operatorname{FAR}(\Gamma) \leq \alpha,
\end{equation}
\noindent where $\alpha$ denotes the desired upper-bound on the false alarm rate.

When $\theta_0$ and $\theta_1$ are known, the Log-Likelihood Ratio (LLR) statistic can be used to recursively compute the CUSUM statistic \cite{Page1954}. As we will discuss next, such a statistic can be used to construct a high-confidence change-point detection method. 
The LLR statistic, $L_t$, and the CUSUM statistic, $W_t$, are updated at each time $t$ as follows:
\begin{equation}
    \label{eq:llr}
    L_t = \log\frac{p_{\theta_1}(S_{t+1}, R_t| S_t, A_t)}{p_{\theta_0}(S_{t+1}, R_t|S_t, A_t)}, 
\end{equation}
\begin{equation}
\label{eq:pagecusum}
    W_t = \max\left(0, W_{t-1} + L_t\right), \ W_0 = 0.
\end{equation}
Importantly, notice that before the change-point $C$ is reached, $\mathbb{E}[L_t] < 0$, which implies that the expected value of $W_t$ is zero. After the change-point $C$ is reached, $\mathbb{E} [L_t] > 0$, and therefore $W_t$ will tend to increase. Higher values of $W_t$, thus, serve as principled statistics reflecting evidence that a change-point has occurred between $\theta_0$ to $\theta_1$. The random time $\Gamma$ when a change-point is estimated to have happened is defined as the first time when the CUSUM metric $W_t$ becomes greater than a detection threshold $h$:
\begin{equation}
\label{eq:tau}
    \Gamma = \min\{t \geq 1 : W_t > h\}.
\end{equation}

In \cite{Lorden1971}, Lorden shows that choosing $h = |\log \alpha|$ ensures that $\operatorname{FAR}(\Gamma) \leq \alpha$. Furthermore, Lai demonstrated that the CUSUM detection time $\Gamma$ is asymptotically optimum \cite{Lai1998} with respect to the problem specified in Eq.~\ref{eq:pollak}. In particular, they showed that the worst expected detection delay (under $h=|\log\alpha|$) respects the following approximation:
\begin{equation}
\label{eq:cusumdelay}
     \Delta_{worst}(\Gamma) \approx \frac{|\log \alpha|}{\mathcal{D}_{KL}(p_{\theta_1}||p_{\theta_0})} \text{ as } \alpha \to 0.
\end{equation}
In Eq.~\ref{eq:cusumdelay}, the denominator $\mathcal{D}_{KL}(p_{\theta_1}||p_{\theta_0}) = \mathbb{E}_{\theta_1} \left[ \log \frac{p_{\theta_1}(S_{t+1}, R_t|S_t, A_t)}{p_{\theta_0}(S_{t+1}, R_t| S_t, A_t)}\right]$ is the Kullback–Leibler divergence under $\theta_1$. Eq.~\ref{eq:cusumdelay} implies that the larger the difference between the distributions $p_{\theta_1}$ and $p_{\theta_0}$, the smaller the expected delay ($\Delta = \Gamma - C$) for detecting a change-point. The above results allow us to construct high-confidence statistics reflecting whether (and when) a context has changed; importantly, they are both accurate and have bounded false alarm rate. 

\section{Model-Based RL Context Detection}

In this section, we introduce an algorithm that iteratively applies a  CUSUM-related procedure to detect context changes under the assumptions discussed in Section \ref{sec:problem}. The algorithm incrementally builds a library of models and policies for tackling arbitrarily different types of contexts; i.e., contexts that may result from quantitatively and qualitatively different underlying causes for non-stationarity---ranging from unpredictable environmental changes (such as random wind) to robot malfunctions. Our method can rapidly deploy previously-constructed policies whenever contexts approximately re-occur, or learn new decision-making strategies whenever novel contexts, with no structural similarities with respect to previously-observed ones, are first encountered. Unlike existing approaches (see Section \ref{sec:related}), our method is capable of \textit{(i)} optimizing policies \textit{online}; \textit{(ii)} without requiring a pre-training phase; \textit{(iii)} based on streams of arbitrarily different contexts, with unknown change-point boundaries; and \textit{(iv)} such that contexts may be drawn from an arbitrary, unknown distribution. 

We now introduce a \textit{high-level} description of our method (\underline{M}odel-\underline{B}ased RL \underline{C}ontext \underline{D}etection, or MBCD). In subsequent subsections, we provide details for each of the method's components. As the agent interacts with a non-stationary environment, context changes are identified via a multivariate variant of CUSUM \cite{Healy1987}, called MCUSUM. MCUSUM-based statistics inherit the same formal properties as those presented in Section \ref{sec:cusum}. In particular, they formally guarantee that MBCD can detect context changes with minimum expected delay, while simultaneously bounding the false alarm rate. As a consequence, MBCD can effectively identify novel environmental dynamics while ensuring, with high probability, that new context-specific policies will only be constructed when necessary. 

As new contexts are identified by this procedure, MBCD updates a mixture model, $M$, composed of a (possibly infinite) ensemble of probabilistic context dynamics predictors, whose purpose is to model the different modes of the distribution over underlying latent MDPs/contexts. New models are added to the ensemble as qualitatively different contexts are first encountered. The mixture model $M$ associates, with each identified context $\mdp_z$, a learned joint distribution $p_{\theta_{z}}$ over next-state dynamics and rewards associated. Let $K$ be the number of context models currently in the mixture. After each agent experience, MBCD identifies the most likely context, $z_t$, by analyzing a set of incrementally-estimated MCUSUM statistics (see Section \ref{sec:changedetection}). Whenever a novel context---one with dynamics that are qualitatively different from those previously-experienced---is observed, a new model is added to the mixture. Context-specific policies, $\pi_{\psi_{z}}$, are trained via a Dyna-style approach \cite{Sutton1990} based on the corresponding learned joint prediction model of $\mdp_z$, $p_{\theta_{z}}$ (see Section~\ref{sec:policyoptimization}). We provide details in the next subsections. Pseudocode for MBCD is shown in Algorithm \ref{alg:mbcd}.

\subsection{Stochastic Mixture Model of Dynamics}
\label{sec:model}

In this paper, we assume that $p_{\theta_z}$, the joint distribution over next-state dynamics and rewards associated with context $\mdp_z$, can be approximated by a multivariate Gaussian distribution. In particular, following recent work on model-based RL \cite{Janner+2019,Chua+2018}, MBCD models the dynamics of a given environment $\mdp_z$, $p_{\theta_z}(S_{t+1}, R_t | S_t, A_t)$, via a bootstrap ensemble of probabilistic neural networks whose outputs parameterize a multivariate Gaussian distribution with diagonal covariance matrix. 
The bootstrapping procedure accounts for epistemic uncertainty (i.e. uncertainty about model parameters), which is crucial when making predictions about the agent's dynamics in regions of the state space where experiences are scarce. For each context $\mdp_z$ identified by MBCD, an ensemble of $N$ stochastic neural networks is instantiated and added to the mixture model $M$. Each network $n$ in the ensemble is parameterized by $\theta^n_z$ and computes a probability distribution, $p_{\theta_{z}^{n}}$, that approximates $p_{\theta_z}$ by predicting the mean and covariance over next-state and rewards conditioned on the current state and action:
\begin{equation}
    p_{\theta_{z}^{n}}(S_{t+1}, R_t | S_t, A_t) = \mathcal{N}(\mu_{\theta_{z}^{n}}(S_t, A_t), \Sigma_{\theta_{z}^{n}}(S_t,A_t)),
\end{equation}
\noindent where $\mu_{\theta_{z}^{n}}(S_t, A_t)$ and $\Sigma_{\theta_{z}^{n}}(S_t,A_t)$ are the network outputs given input $(S_t, A_t)$.
To simplify notation, let $\xt = (S_t, A_t) \in \mathbb{R}^{\dim(\mathcal{S}) + \dim(\mathcal{A})}$ and $\yt = (S_{t+1}, R_t) \in \mathbb{R}^{\dim(\mathcal{S})+1}$.
We follow Lakshminarayanan et al. \cite{Lakshminarayanan+2017} and model the ensemble prediction as a Gaussian distribution whose mean and covariance are computed based on the mean and covariances of each component of the ensemble. In particular, the ensemble predictive model, $\hat{p}_{\theta_z}$, associated with a given context $\mdp_z$, is defined as:
\begin{equation}
    \hat{p}_{\theta_z}(\yt|\xt) = \mathcal{N}(\mu^*_z(\xt), \Sigma^*_z(\xt)),
\end{equation}    
\noindent where
\begin{gather} 
    \mu^*_z(\xt) = N^{-1} \sum_{n=1}^{N} (\mu_{\theta^n_z}(\xt)), \mbox{ and} \\ 
    \Sigma^*_z(\xt) = N^{-1} \sum_{n=1}^{N} \big( \mbox{diag}(\Sigma_{\theta^n_z}(\xt)) + \mu^2_{\theta^n_z}(\xt) \big) - \mu
^2_*(\xt).
\end{gather}

MBCD stores all experiences collected while in context $\mdp_z$ in a buffer $D_{z}$. After every $F$ steps, it uses data in  $D_{z}$ to update the ensemble model $\hat{p}_{\theta_{z}}$ by minimizing the negative log prediction likelihood loss function, $\mathcal{J}_p(\theta,D) = \mathbb{E}_{(s_t, a_t, r_t, s_{t+1}) \sim D}[-\log p_\theta(s_{t+1}, r_t | s_t, a_t)]$.

\subsection{Online Context Change-Point Detection}
\label{sec:changedetection}

As previously discussed, MBCD employs a multivariate variant of CUSUM, MCUSUM, to detect context changes with high confidence. Healy demonstrated that, when detecting shifts in the mean of a multivariate Gaussian distribution, MCUSUM inherits all theoretical optimality guarantees possessed by the univariate CUSUM procedure \cite{Healy1987}. Furthermore, he also proved that the detection delay is independent of the dimensionality of the data.

In the particular case where the dynamics of each context are modeled as multivariate Gaussians, the LLR statistic can be computed as follows. To simplify notation, let $\mu_0 = \mu_{\theta_0}(S_t,A_t)$ and $\Sigma_0 = \Sigma_{\theta_0}(S_t,A_t)$. It is then possible to show that the LLR statistic, $L_t$, between distributions $p_{\theta_1}(\yt|\xt)$ and $p_{\theta_0}(\yt|\xt)$, is given by:
\begin{align}
    L_t &= \log\frac{(2\pi)^{-\frac{d}{2}} |\Sigma_1|^{-\frac{1}{2}} \exp\{-0.5 (\yt-\mu_1) \Sigma_1^{-1} (\yt-\mu_1)\}} {(2\pi)^{-\frac{d}{2}}|\Sigma_0|^{-\frac{1}{2}} \exp\{-0.5 (\yt-\mu_0) \Sigma_0^{-1} (\yt-\mu_0)\}}
\end{align}
\noindent where $d$ is the dimensionality of the multivariate Gaussian. At each time step $t$, MBCD uses $L_t$ to compute MCUSUM statistics $W_{k,t}$ for each known context $k$, plus an additional statistic $W_{\small{\mbox{new}},t}$, used to infer, with high probability, whether a novel context has been first encountered:
\begin{equation}
\begin{aligned}
\label{eq:cusum}
    W_{k,t} \leftarrow \max \left( 0, W_{k,t-1} + \log{\frac{p_{\theta_{k}}(\yt|\xt)}{p_{\theta_{z_t}}(\yt|\xt)}}\right),\ \forall k \in [1,K] \cup [\mbox{new}].
\end{aligned}
\end{equation}
\noindent Here, $W_{\small{\mbox{new}},t}$ can be seen as evidence that a previously-unseen context has been first encountered, based on whether the likelihood of all known contexts $k \in [1,K]$ is smaller than  $p_{\theta_{\small{\mbox{new}}}}$, where
\begin{equation}
\label{eq:pthetanew}
    p_{\theta_{\small{\mbox{new}}}}(\yt  \,|\,\xt) = \mathcal{N}(\yHatt,\, \Sigma_{\theta_{z_t}}(\xt)),
\end{equation}
\noindent and 
\begin{equation}
\yHatt = \yt + \delta \, \mbox{diag}( \Sigma_{\theta_{z_t}}(\xt)).
\end{equation}
\noindent Intuitively,  $W_{\small{\mbox{new}},t}$ indicates whether none of the known contexts is likely to have generated the observed transitions. In Eq.~\ref{eq:pthetanew}, $p_{\theta_{\small{\mbox{new}}}}$ is the likelihood of a new context under the alternative hypothesis that the true observation $\yt$ is $\delta$ standard deviations away from the true observation $\yt$. In particular, $\delta$ indicates the minimum meaningful change in the distribution's mean that we are interested in detecting. 
Given updated statistics $W_{k,t}$, the most likely current context, $z_t$ (which may or may not have changed) can then be identified as:
\begin{equation}
    \label{eq:zt}
    z_t \leftarrow 
    \begin{cases}
        \argmax_k W_{k,t}, & \text{if } \exists k \in [1,K] \cup [\mbox{new}] \mbox{ s.t. } W_{k,t} > h, \\
        z_{t-1}, & \text{otherwise}.
    \end{cases}
\end{equation}
\noindent If no alternative contexts are more likely to have generated the observations collected up to time $t$, no context change is detected and $z_t = z_{t-1}$.

Notice that in Eq.~\ref{eq:cusum}, models $p_{\theta_k}$ are assumed to be known \textit{a priori}. In our setting, these models are estimated based on samples. 
MCUSUM has been studied in scenarios where the parameters of the distribution are known only approximately \cite{Mahmoud+2013}.
In our work, we address this challenge by computing change-point detection statistics only after a small warm-up period within which the agent is allowed to operate in a given context.
In particular, and similarly to Sekar et al. \cite{Sekar+2020}, we define the warm-up period by using ensemble disagreement as a proxy to quantify the system's uncertainty regarding the current distributions. Assuming a warm-up period where the agent is allowed to operate within each newly-encountered context is a common assumption in the area \cite{Nagabandi+2019a,Nagabandi+2019b,Rakelly+2019}. In fact, it is a \textit{necessary} assumption: if contexts are allowed to change arbitrarily fast, adversarial settings can be constructed where all methods for dealing with non-stationary scenarios fail.
    
Finally, notice that a key element of Eq.~\ref{eq:zt} is the threshold $h$, against which each $W_{k,t}$ is compared in order to check if a context change has occurred. Different methods have been proposed to set $h$ \cite{Sahki+2020}. Here, we take a conservative approach. As discussed in Section \ref{sec:cusum}, setting $h = |\log \alpha|$ ensures that $FAR(\Gamma) \leq \alpha$. In this work, we set $h$ by considering negligible values of $\alpha$; e.g. $h=100$ if $\alpha \approx 10^{-43}$. In our experiments, we observed that detection delays remain low even for very conservative values of $h$.

\subsection{Policy Optimization}
\label{sec:policyoptimization}

Since MBCD estimates dynamics models for each context, it is natural to exploit such models to accelerate policy learning by deploying model-based RL algorithms.
MBCD learns context-specific policies via a Dyna-style approach \cite{Sutton1990}. In particular, at every time step $t$ during which the model $p_{\theta_{z_t}}$ is trained, $L$ 1-step simulated rollouts are sampled using $\pi_{\psi_{z_t}}$. Each rollout starts from a random state drawn from $D_{z_t}$. All rollouts are stored in a buffer, $D_{model}$.\footnote{Notice that, in Algorithm \ref{alg:mbcd}, $D_{model}$ is cleared every time a context change is detected in order to avoid negative transfer from experiences drawn from previous contexts.} Notice that this is similar to the procedure used by the Model-Based Policy Optimization (MBPO) algorithm \cite{Janner+2019}. Each context-specific policy $\pi_{\psi_{z_t}}$ is  optimized by taking into account both real experiences (stored in $D_{z_t}$) and simulated experiences (stored in $D_{model}$). Policy optimization is performed using the Soft Actor-Critic (SAC) algorithm \cite{Haarnoja+2018}. SAC alternates between a policy evaluation step, which estimates $q_\pi(s,a) = \mathbb{E} [\sum_{t=0}^{\infty}\gamma^tR_t| S_t=s,A_t=a, \pi]$ using the Bellman backup operator, and a policy improvement step, which optimizes the policy $\pi$ by minimizing the expected KL-divergence between the current policy and the exponential of a soft Q-function \cite{Haarnoja+2018}. Optimizing the policy, then, corresponds to minimizing the following loss function:
\begin{equation}
\mathcal{J}_\pi(\psi,D) = \mathbb{E}_{s_t \sim D} \left[ \mathbb{E}_{a_t \sim \pi_{\psi}} (\beta \log(\pi_\psi(a_t|s_t)) - q_{\pi_\psi}(s_t,a_t)\right].
\end{equation}

\begin{algorithm}[ht]
\caption{Model-Based RL Context Detection}
\label{alg:mbcd}

\DontPrintSemicolon
\SetKwInOut{Input}{Input}

\Input{Non-stationary environment $E$, threshold $h$.}
 $z_0 \leftarrow 1$; $K \leftarrow 1$; $W_{z_0,0} \leftarrow 0$; $W_{\small{\mbox{new}},0} \leftarrow 0$\\
 Initialize model $p_{\theta_{z_0}}$, policy $\pi_{\psi_{z_0}}$, datasets $D_{z_0}$ and $D_{model}$\;
 $M \leftarrow \{p_{\theta_{z_0}}\}$\;

\For{$t = 0 ... \infty$}{

Execute action $a_t \sim \pi_{\psi_{z_t}}(s_t)$, observe $s_{t+1}$, $r_t$\;

Update MCUSUM statistics $W_{k,t}, \forall k$, with Eq.~\ref{eq:cusum}\;

Update $z_t$ with Eq.~\ref{eq:zt}\;

\If(\tcp*[h]{Context changed}){$z_t \neq z_{t-1}$}{ 
    Reset MCUSUM statistics\;
    $D_{model} =\{ \}$\;
    \If(\tcp*[h]{New context detected}){$z_t =$ new}{
          $K \leftarrow K + 1$; $z_t \leftarrow K$\;
          Initialize $p_{\theta_{z_t}}$ and $D_{z_t}$\;
          Let $\pi_{\psi_{z_t}} \leftarrow \pi_{\psi_{z_{t-1}}}$\;
          $M \leftarrow M \cup \{p_{\theta_{z_t}}$\}\; 
    }
}

$D_{z_t} \leftarrow D_{z_t} \cup \{(s_t,a_t,r_t,s_{t+1})\}$\;

\If{$t \bmod F = 0$}{
    $\theta_{z_t} \leftarrow \theta_{z_t} - \lambda_p \nabla \mathcal{J}_{p}(\theta_{z_t}, D_{z_t})$ \tcp*{Update model}

    \For{$L$ simulated 1-step rollouts}{
      Sample state $s_i$ from tuples in $D_{z_t}$\;
      $a_i \sim \pi_{\psi_{z_t}}(\cdot|s_i)$\;
      $(s'_i, r_i) \sim p_{\theta_{z_t}}(\cdot|s_i,a_i)$\;
      $D_{model} \leftarrow D_{model} \cup (s_i, a_i, r_i, s'_i)$\;
    }
}

$\psi_{z_t} \leftarrow \psi_{z_t} - \lambda_\pi \nabla \mathcal{J}_{\pi}(\psi_{z_t}, D_{model} \cup D_{z_t})$ \tcp*{Update policy}
}

\end{algorithm}

\section{Experiments}

We evaluate MBCD in challenging continuous-state, continuous-action non-stationary environments, where the non-stationarity may result from qualitatively different reasons---ranging from abrupt changes to the system's dynamics (such as changes to the configuration of the agent's workspace); external latent environmental factors that impact the distribution over next states (such as random wind); robot malfunctions; and changes to the agent's goals (its reward function). We compare MBCD both against state-of-the-art RL algorithms and against state-of-the-art meta-learning methods specifically tailored to deal with non-stationary environments.

We investigate the performance of MBCD in two settings: \textit{(i)} a setting that satisfies all standard requirements made by meta-learning algorithms; in particular, that all contexts in ${M_z}$ are structurally similar, and that the agent is tested on a distribution of contexts that matches the training distribution; and \textit{(ii)} a more general setting where such requirements are not be satisfied: contexts may differ arbitrarily, and future contexts experienced by the agent may be off-distribution with respect to those sampled during the training phase. We show that our method outperforms both standard state-of-the-art RL algorithms and also specialized meta-learning algorithms in both settings.

\begin{figure}
    \centering
    \includegraphics[width=0.3\textwidth]{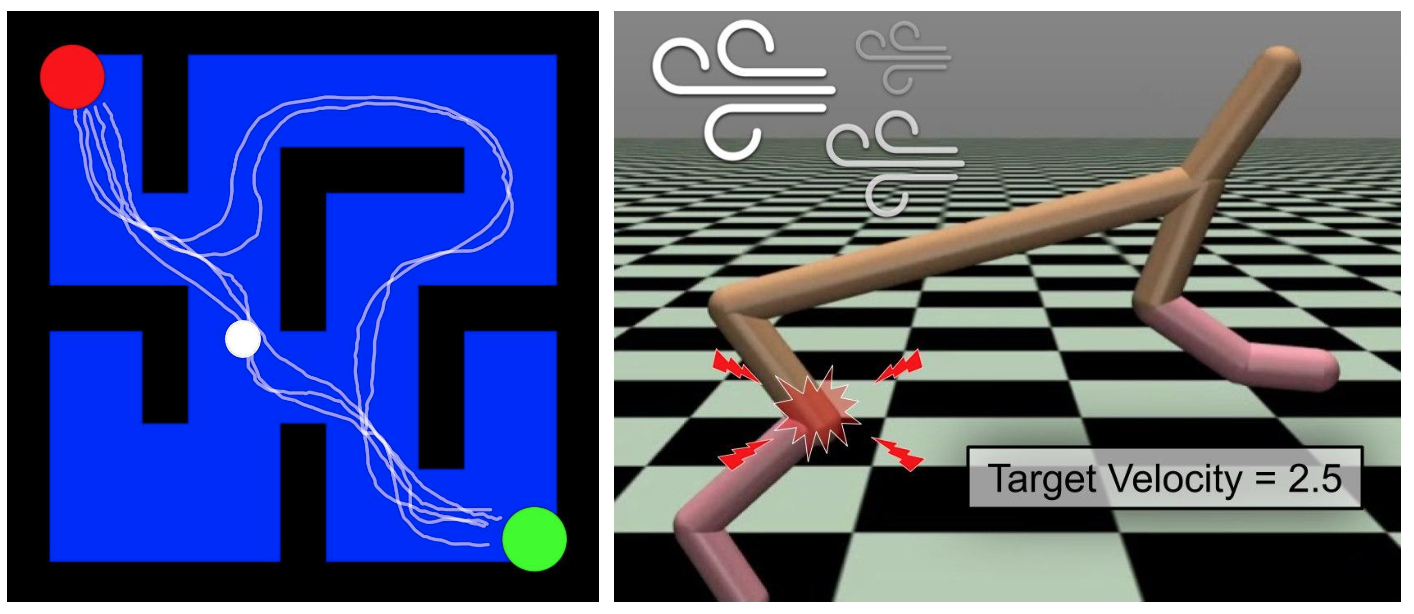}
    \Description{(a) Non-Stationary Continuous Particle Maze; (b) Half-Cheetah in a Non-Stationary World.}
    \caption{(a) Non-Stationary Continuous Particle Maze; (b) Half-Cheetah in a Non-Stationary World.}
    \label{fig:environments}
\end{figure}

In what follows, we evaluate MBCD in two domains with qualitatively different non-stationary characteristics (see Fig.~\ref{fig:environments}):

\paragraph{Non-Stationary Continuous Particle Maze}
This domain simulates a family of two-dimensional continuous mazes where a particle must reach a non-observable goal location. The reward function corresponds to the Euclidean distance between the particle and the goal location. Non-stationarity is introduced by either changing the location of walls or by randomly changing the latent goal location.
\paragraph{Half-Cheetah in a Non-Stationary World}
This domain consists of a simulation of the high-dimensional \textit{Half-Cheetah} robot \cite{Todorov+2012}. The agent's goal is to move forward while reaching a particular target-velocity and minimizing control costs. We introduce three sources of non-stationarity:
\begin{enumerate}
    \item \textbf{random wind}: an external latent horizontal force, opposite to the agent's movement direction, is applied;
    \item \textbf{joint malfunction}: either the torque applied to a joint of the robot has its polarity/sign changed; or a joint is randomly disabled;
    \item \textbf{target velocity}: the target velocity of the robot is sampled from the interval 1.5 to 2.5, causing a non-stationary change to the agent's reward function.
\end{enumerate}

\begin{figure}[ht!]
\centering
\begin{subfigure}[t]{\columnwidth}
\centering
\includegraphics[width=\textwidth]{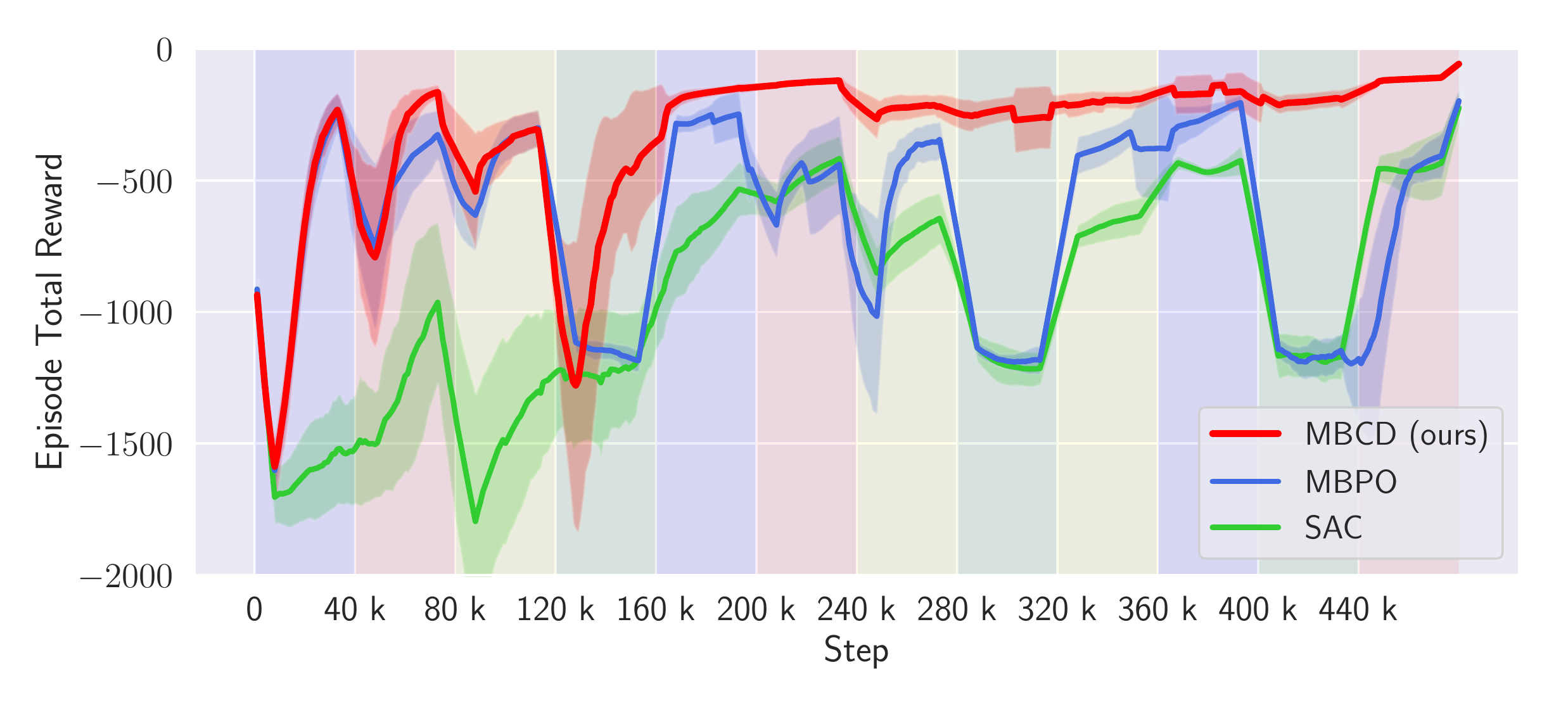}
\Description{Total reward achieved by different methods (MBCD, MBPO, and SAC) as contexts change.}
\caption{Total reward achieved by different methods (MBCD, MBPO, and SAC) as contexts change.}
\label{fig:exp1reward}
\end{subfigure}
\begin{subfigure}[t]{\columnwidth}
\centering
\includegraphics[width=\textwidth]{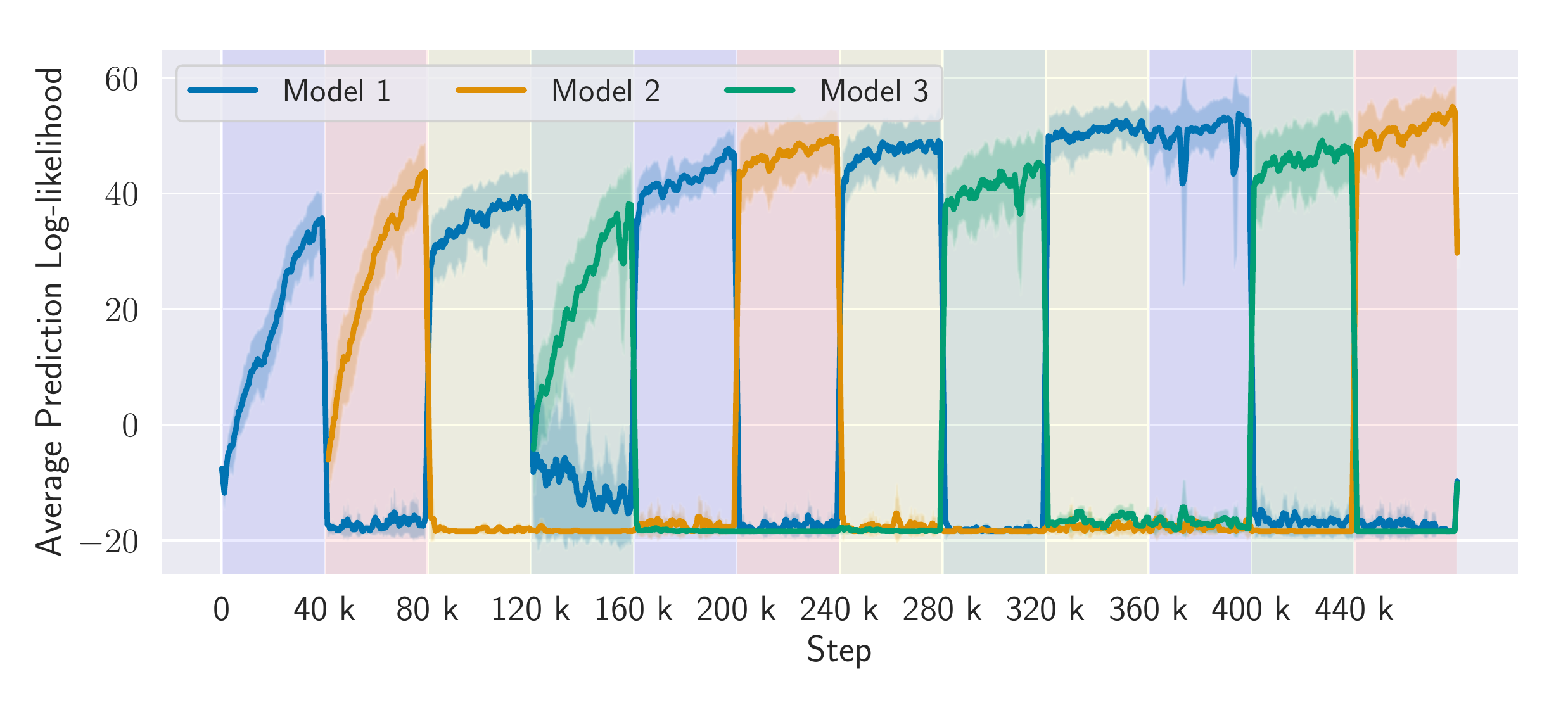}
\Description{Log-likelihood of the predictions made by each context model learned by MBCD as contexts change.}
\caption{Log-likelihood of the predictions made by each context model learned by MBCD as contexts change.}
\label{fig:exp1prob}
\end{subfigure}
\begin{subfigure}[t]{\columnwidth}
\centering
\includegraphics[width=\textwidth]{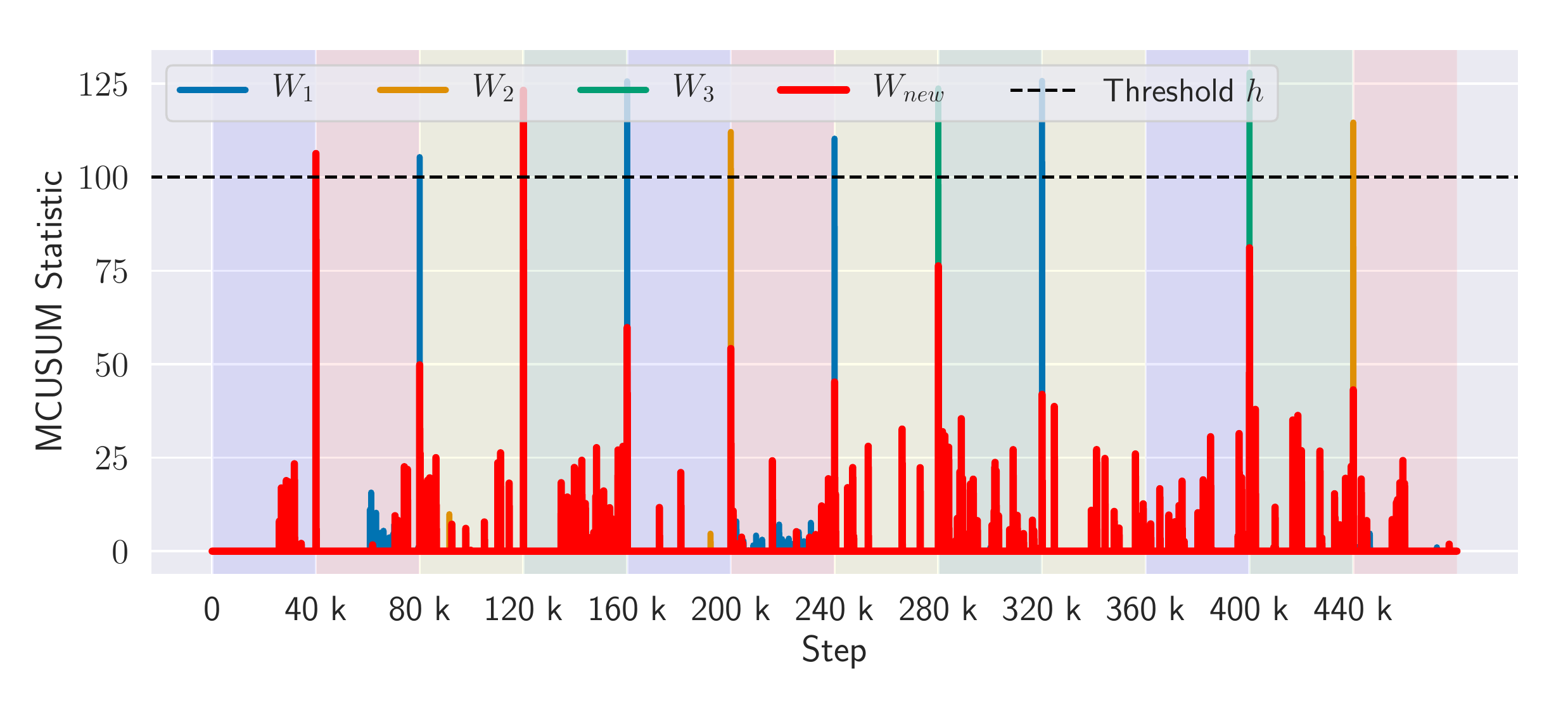}
\Description{Time evolution, as contexts change, of the MCUSUM  statistics, $W_k$, for each model $k$. A context change is detected online whenever one of the statistics crosses the threshold $h$.}
\caption{Time evolution, as contexts change, of the MCUSUM  statistics, $W_k$, for each model $k$. A context change is detected online whenever one of the statistics crosses the threshold $h$.}
\label{fig:exp1cusum}
\end{subfigure}
\Description{Evaluation of MBCD on the non-stationary \textit{Half-Cheetah} domain. Colored shaded areas represent different contexts: \textit{(blue)} default context; \textit{(red)} joint malfunction; \textit{(yellow)} wind; \textit{(green)} novel target velocity.}
\caption{Evaluation of MBCD on the non-stationary \textit{Half-Cheetah} domain. Colored shaded areas represent different contexts: \textit{(blue)} default context; \textit{(red)} joint malfunction; \textit{(yellow)} wind; \textit{(green)} novel target velocity.}
\label{fig:exp1}
\end{figure}

We first evaluate our method on the non-stationary Half-Cheetah domain and compare it with two state-of-the-art RL algorithms: MBPO \cite{Janner+2019} and SAC \cite{Haarnoja+2018}. In our setting, MBPO can be seen as a particular case of our algorithm, where a single dynamics model and policy are tasked with optimizing behavior under changing contexts. SAC works similarly to MBPO but does not perform Dyna-style planning steps using a learned dynamics model.

Fig.~\ref{fig:exp1reward} shows the total reward achieved by different methods (ours, MBPO, SAC) as contexts change. Colored shaded areas depict different contexts, as discussed in the figure's caption. Notice that our method and MBPO have similar performances when interacting for the first time with the first three random contexts. In particular, both MBCD and MBPO's performances temporarily drop when a novel context is encountered for the first time. MBCD's performance drops because it instantiates a new dynamics model for the newly encountered context, while MBPO's performance drops because it undergoes negative transfer. SAC, which is model-free, never manages to achieve reasonable performance during the duration of each context, due to sample inefficiency. However, as the agent encounters contexts with structural similarities with respect to previously-encounters ones (around time step 160k), MBCD's performance becomes near-optimal: it rapidly identifies whenever a context change has occurred and deploys an appropriate policy.\footnote{Notice that MBCD modeled the wind context (yellow area) using the same model as the default context (blue area). This is because wind did not introduce a significant change to the MDPs state transition function. Consequently, MBCD automatically inferred that a single policy could perform well in both contexts and operated without significant reward loss in the long term---see, e.g., time steps 320k to 400k.} MBPO and SAC, on the other hand, suffer from negative transfer due to learning average policies or dynamics models. They are also subject to catastrophic forgetting and do not reuse previously-acquired, context-specific knowledge.

Fig.~\ref{fig:exp1prob} and Fig.~\ref{fig:exp1cusum} allow us to observe the inner workings of MBCD and understand the reasons that underlie its performance. Fig.~\ref{fig:exp1prob} shows the log-likelihood of the predictions made by each context model learned by MBCD as contexts change. Notice that all context changes in this experiment---even those caused by qualitatively different sources of non-stationarity---are detected with minimal delay. As contexts change, MBCD rapidly detects each change and instantiates new specialized joint prediction models for each context. Furthermore, when structurally similar contexts are re-encountered (e.g., at time steps 160k and timestep 360k), MBCD successfully recognizes that previously-learned models may be redeployed and avoids having to relearn context-specific dynamics or policies. Fig.~\ref{fig:exp1cusum} shows the time evolution, as contexts change, of the MCUSUM  statistics, $W_k$, for each model $k$ in the ensemble. A context change is detected, online, whenever one of the statistics crosses the threshold $h$. Notice that MCUSUM statistics grow rapidly and cross the threshold almost instantaneously---only a few steps after a context change. This empirically confirms the minimum-delay guarantees provided by our change-point detection method.

\begin{figure}
    \centering
    \includegraphics[width=0.4\textwidth]{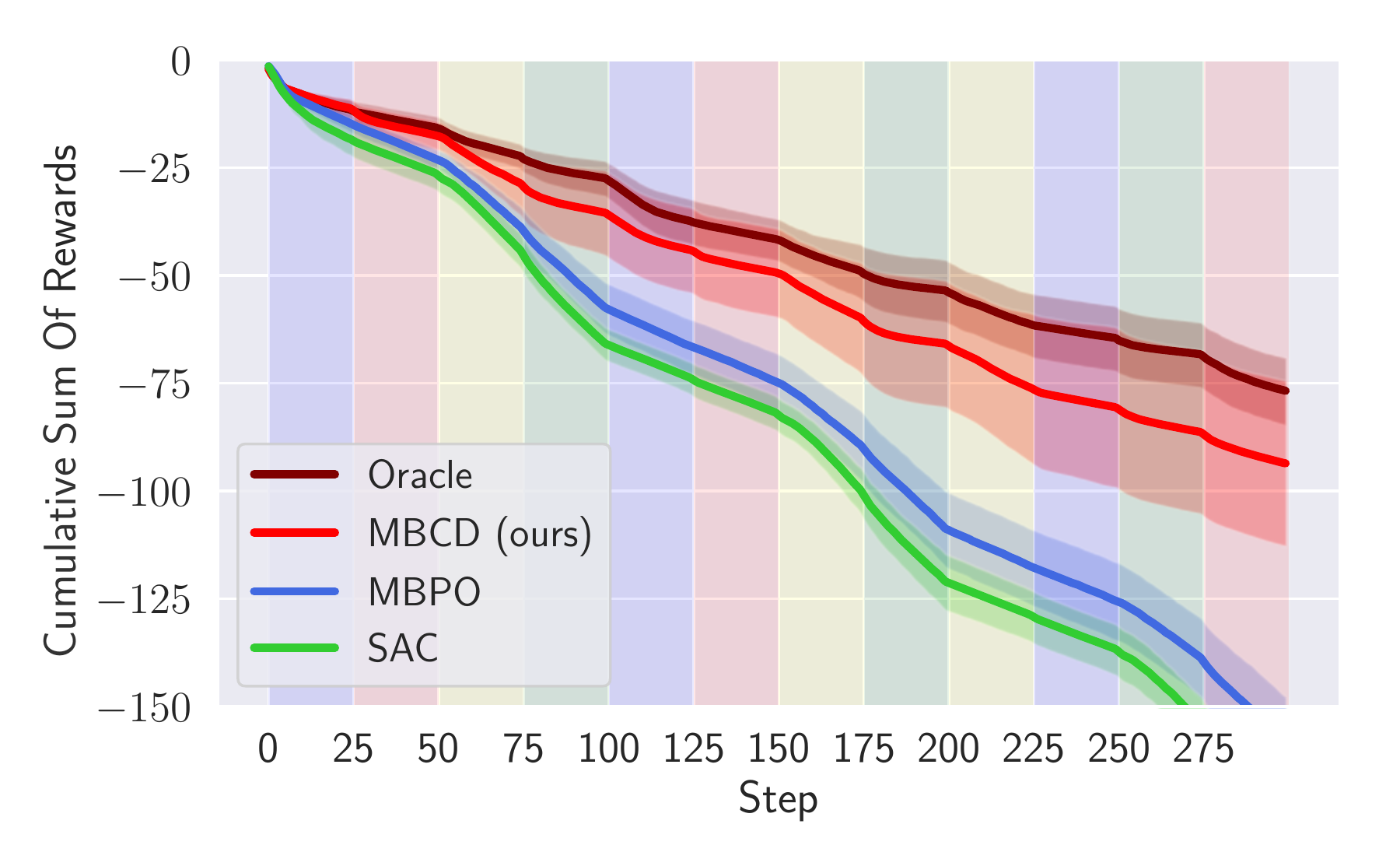}
    \Description{Regret of different methods (MBCD, MBPO, SAC) in the non-stationary \textit{Half-Cheetah} domain, as contexts change rapidly, compared to an Oracle algorithm.}
    \caption{Regret of different methods (MBCD, MBPO, SAC) in the non-stationary \textit{Half-Cheetah} domain, as contexts change rapidly, compared to an Oracle algorithm.}
    \label{fig:exporacle}
\end{figure}

We now evaluate MBCD's ability to rapidly detect context changes after an initial training period. Fig.~\ref{fig:exporacle} shows the cumulative sum of rewards achieved by MBCD, MBPO, SAC, and by an Oracle algorithm that is initialized with optimal policies for all contexts and that detects context changes with zero delay. This is a challenging setting where contexts change very rapidly---after only 25 steps. Notice that our algorithm closely matches the performance of the zero-delay Oracle, thus empirically confirming its ability to minimize regret (Eq.~\ref{eq:regret}).

\begin{figure}
    \centering
    \includegraphics[width=0.4\textwidth]{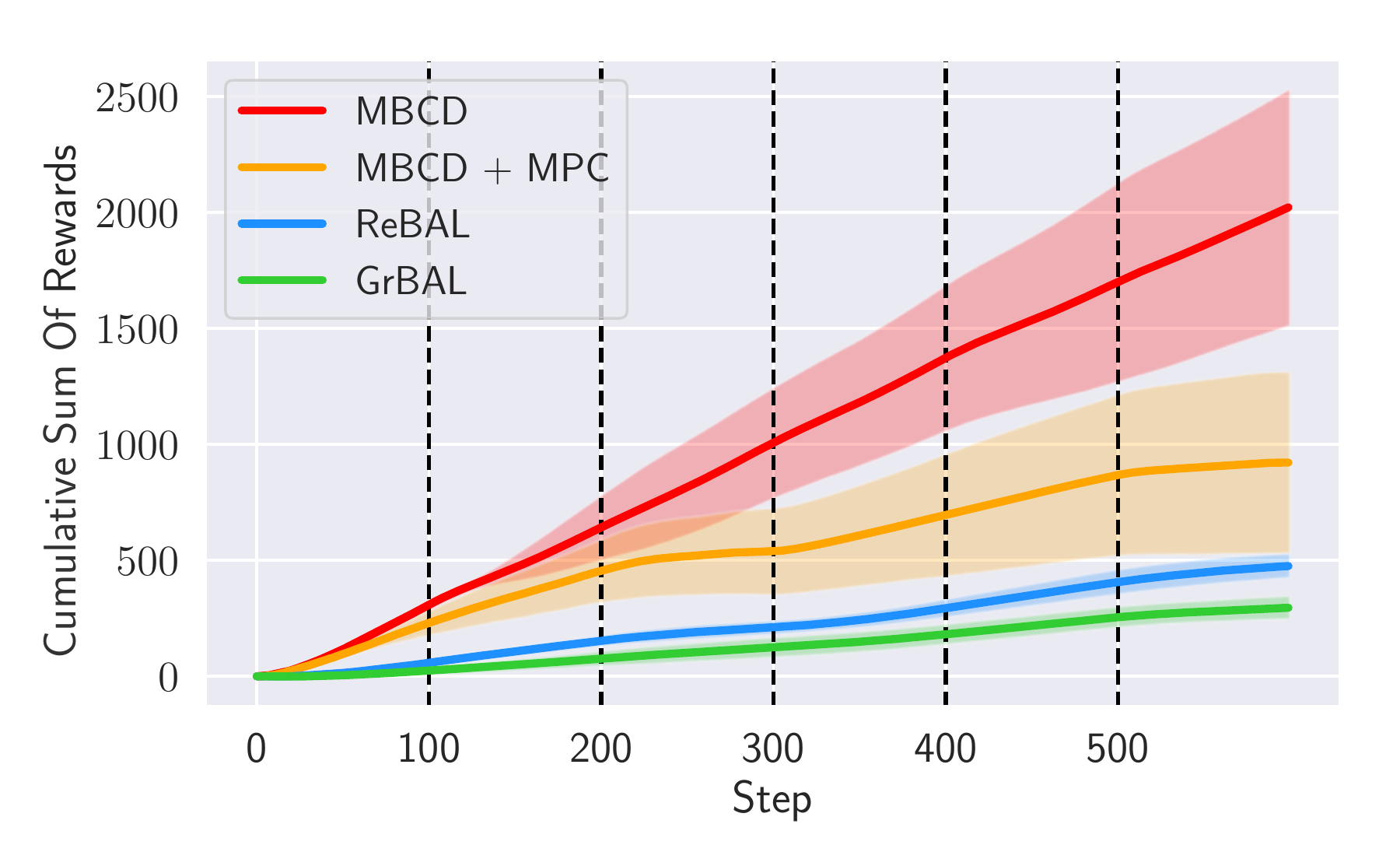}
    \Description{Performance of MBCD and meta-learning methods (after a pre-training phase) in the \textit{Half-Cheetah} domain with non-stationary malfunctions that disable random joints. Vertical dashed lines indicate context changes.}
    \caption{Performance of MBCD and meta-learning methods (after a pre-training phase) in the \textit{Half-Cheetah} domain with non-stationary malfunctions that disable random joints. Vertical dashed lines indicate context changes.}
    \label{fig:expmeta}
\end{figure}

Next, we analyze how MBCD performs when compared with state-of-the-art meta-learning methods specifically tailored to deal with non-stationary environments: Gradient-Based Adaptive Learner\footnote{We used the authors' implementation of the method, publicly available at \url{https://github.com/iclavera/learning_to_adapt}.\label{footnotemeta}} (GrBAL) \cite{Nagabandi+2019a}  and Recurrence-Based Adaptive Learner\textsuperscript{\ref{footnotemeta}} (ReBAL) \cite{Nagabandi+2019a}. 
GrBAL employs the Model-Agnostic Meta-Learning (MAML) \cite{Finn+2017} method to learn the parameters of a meta-learning prior over the dynamics model, given a set of training contexts. This prior is constructed so that it serves as a good initial model for any new contexts that the agent encounters after a pre-training phase. After training, such a meta-learned dynamics model is capable of quickly adapting to a current task's dynamics by taking only a few gradient steps. ReBAL works similarly to GrBAL, but instead of taking gradient steps to adapt a prior model to novel contexts, it uses a recurrent neural network that learns its own update rule (vs. a gradient update rule) through its hidden state. Both ReBAL and GrBAL use Model-Predictive Control (MPC) \cite{Garcia+1989} for selecting actions by planning for a certain horizon using the learned dynamics model.

Fig.~\ref{fig:expmeta} compares the adaptation performance of MBCD and the meta-learning methods in a non-stationary setting where (inspired by \cite{Nagabandi+2019a}) random joints of the Half-Cheetah robot are disabled after every 100 time steps. In this experiment we compare MBCD, ReBAL, GrBAL, and also (for fairness) a variant of MBCD that chooses actions using MPC instead of SAC. All implementations of MPC make use of the cross entropy method (CEM) \cite{Botev+2013} to accelerate action selection. All methods are allowed to interact with each randomly-sampled context during a training phase comprising 60000 time steps.
Although the meta-learning methods have lower-variance, their meta-prior models do not perform as well as the MBCD context-specific dynamics models and policies. We also observe that when MBCD uses parameterized policies, learned through Dyna-style planning, it performs better than MBCD coupled with MPC.

\begin{figure}
    \centering
    \includegraphics[width=0.4\textwidth]{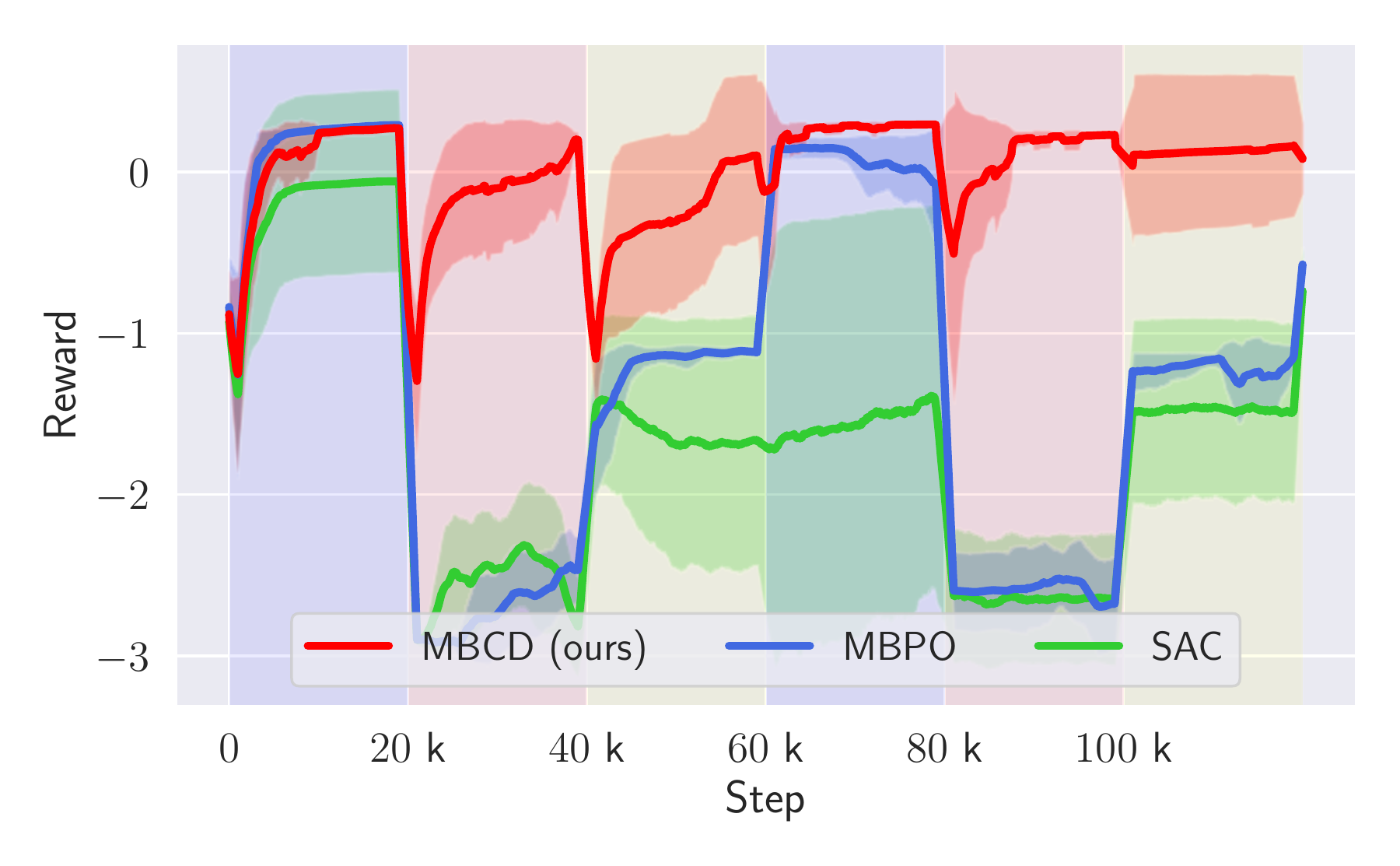}
    \Description{Rewards in the non-stationary continuous particle maze domain. Shaded colored areas indicate different contexts: \textit{(blue)} default maze; \textit{(red)} maze with non-stationary wall positions; \textit{(yellow)} non-stationary target positions.}
    \caption{Rewards in the non-stationary continuous particle maze domain. Shaded colored areas indicate different contexts: \textit{(blue)} default maze; \textit{(red)} maze with non-stationary wall positions; \textit{(yellow)} non-stationary target positions.}
    \label{fig:expmaze}
\end{figure}

We now evaluate MBCD in a setting where contexts may differ arbitrarily and where future contexts may be off-distribution with respect to those sampled during the training phase. To do this, we compare MBCD, MBPO, SAC, ReBAL, and GrBAL, in the non-stationary continuous particle maze domain, where the sources of non-stationarity are as discussed earlier.
\balance
Fig.~\ref{fig:expmaze} compares MBCD, MBPO, and SAC in the fully-online setting---no pre-training phase is allowed. Notice that, even in this relatively simple scenario, state-of-the-art RL algorithms may fail if the underlying state transition or reward function changes drastically. MBCD's performance, by contrast, remains approximately constant (and high) as contexts change.
\begin{figure}
    \centering
    \includegraphics[width=0.4\textwidth]{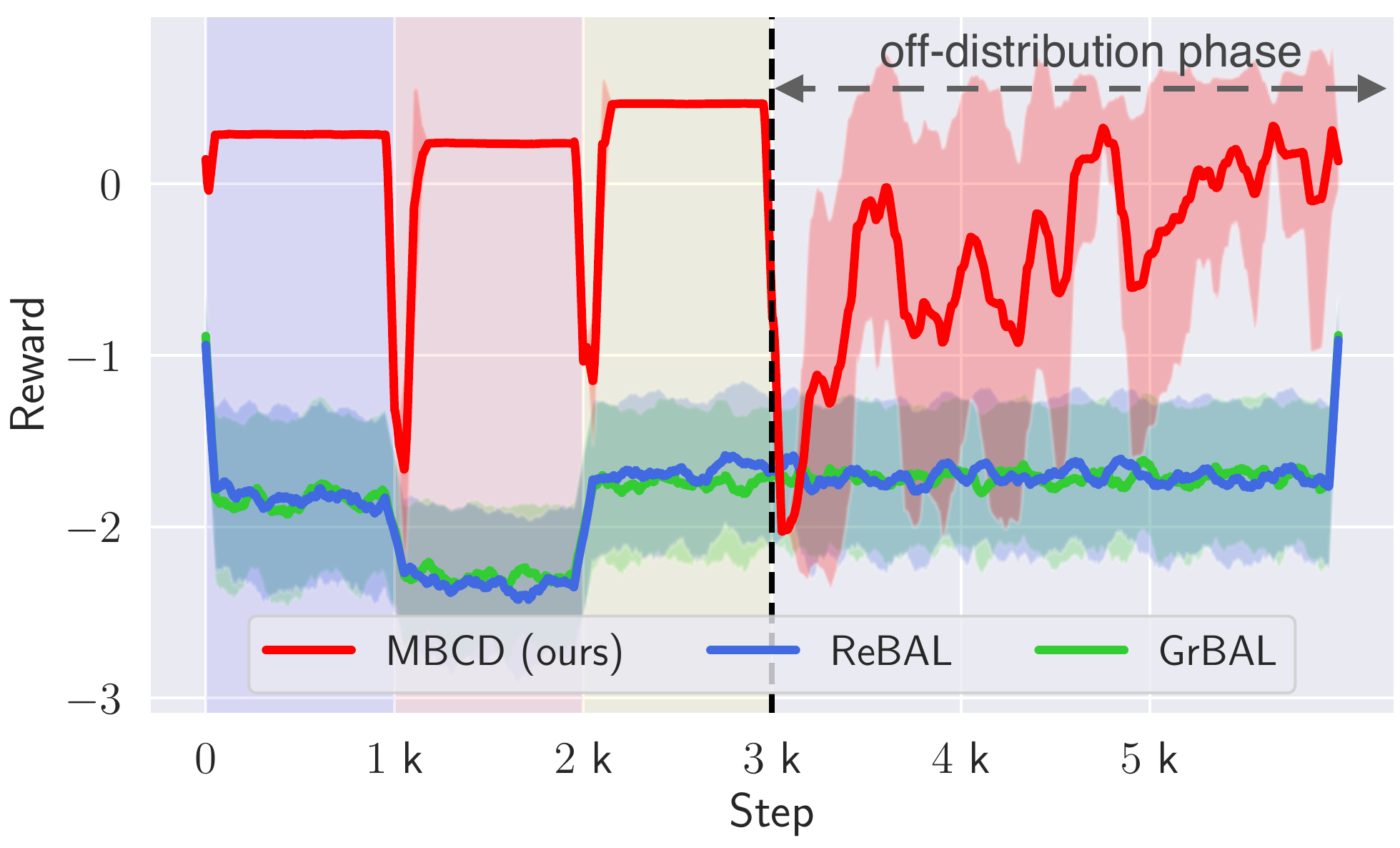}
    \Description{Rewards in the non-stationary maze domain. We introduce a phase with off-distribution contexts---contexts unlike those observed during pre-training.}
    \caption{Rewards in the non-stationary maze domain. We introduce a phase with off-distribution contexts---contexts unlike those observed during pre-training.}
    \label{fig:expmazemeta}
\end{figure}
In Fig.~\ref{fig:expmazemeta}, we compare MBCD with meta-learning methods after a phase of pre-training. All contexts observed up to time step $3k$ are on-distribution: they are similar to those experienced during training. MBCD outperforms ReBAL and GrBAL because the meta-learning approaches construct models that try to average characteristics of structurally different contexts. At time $3k$, we initiate a phase with off-distribution contexts---contexts unlike those observed during training. When this occurs, MBCD faces a short adaption period and degrades gracefully, while meta-learning techniques perform poorly. This emphasizes MBCD's advantages over meta-learning models, both when a pre-training phase is not allowed/possible (e.g., Fig~\ref{fig:expmaze}) and also when testing contexts arise from a distribution different from the training distribution.

\section{Related Work}
\label{sec:related}

Dealing with non-stationary environments in RL via context change-point detection has been studied in discrete state and action spaces settings. da Silva et al. \cite{Silva+2006icml} introduced Reinforcement Learning Context Detection (RLCD). RLCD is a model-based algorithm that estimates the prediction quality of different models, and instantiates new ones when none of the existing models performs well. Even though it does not require a pre-training phase (like meta-learning algorithms), it is only applicable to purely discrete settings.
Hadoux et al. introduced an extension of RLCD that uses a CUSUM-based method to perform change-point detection \cite{Hadoux+2014}. Unlike our method, however, it is only applicable to discrete state settings. Banerjee et al. proposed a two-threshold switching policy based on KL divergence between transition models in order to rapidly detect context changes \cite{Banerjee+2017}. This is a principled method but---unlike MBCD---requires prior knowledge of the dynamics model of all contexts.

In \cite{Nagabandi+2019a} and \cite{Nagabandi+2019b}, meta-learning algorithms are used to train a prior over dynamics models that can, when combined with recent data, be rapidly adapted to novel contexts.
These methods, unlike MBCD, were designed to tackle settings where the non-stationarity solely results from changes to the dynamics, but not to the agent's goals/reward function. Reward functions are assumed to be known \textit{a priori}. These methods also require an explicit pre-training train phase, prior to deployment, and assume that the distribution of training and testing contexts is the same. Our method, by contrast, is better suited to continual online settings where a pre-training phase is not possible, and where the agent is tasked with dealing with streams of arbitrarily different contexts with unknown boundaries.

\section{Conclusion}

We introduced a model-based reinforcement learning algorithm (MBCD) that learns efficiently in non-stationary settings with continuous states and actions. It makes use of high-confidence change-point detection statistics to detect context changes with minimum delay, while bounding the rate of false alarm. It is capable of optimizing policies online, without requiring a pre-training phase, even when faced with streams of arbitrarily different contexts drawn from unknown distributions. We empirically show that it outperforms state-of-the-art (model-free and model-based) RL algorithms, and that it outperforms state-of-the-art meta-learning methods specially designed to deal with non-stationarity---in particular, if the agent is faced with MDPs that are off-distribution with respect to the set of training contexts. As future work, we would like to extend our method so that it can actively transfer knowledge of learned policies between contexts. This research direction suggests that our method may be combined with meta-learning techniques that operate over policies, instead of over dynamics models.



\begin{acks}
This study was financed in part by the Coordenação de Aperfeiçoamento de
Pessoal de Nível Superior - Brasil (CAPES) - Finance Code 001. Ana Bazzan was partially supported by CNPq under grant no. 307215/2017-2.
\end{acks}



\bibliographystyle{ACM-Reference-Format}



\end{document}


\maketitle

\section{Environments}

\begin{figure}[h]
    \centering
    \includegraphics[width=0.6\linewidth]{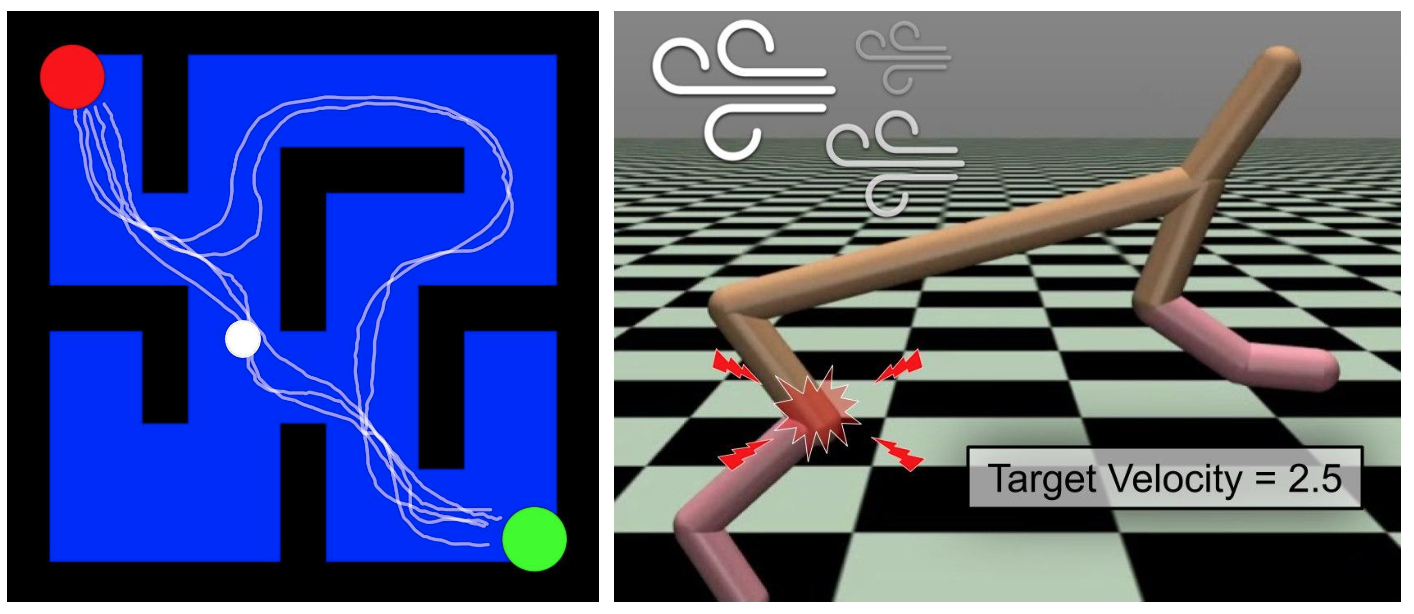}
    \caption{(a) Non-Stationary Continuous Particle Maze; (b) Half-Cheetah in a Non-Stationary World.}
    \label{fig:environments}
\end{figure}

We evaluate MBCD in two domains with qualitatively different non-stationary characteristics (see Fig.~\ref{fig:environments}).

\subsection{Non-Stationary Continuous Particle Maze}
This domain simulates a family of two-dimensional continuous mazes where a particle must reach a non-observable goal location.
\begin{itemize}
    \item Observation. $S_t = [x_{pos}, y_{pos}]$. The observed state consists of the particle coordinates in the continuous grid.
    \item Action. $A_t \in [-1.0,1.0]^2$. Represents the direction of movement along the two dimensions.
    \item Reward. $R_t = - ||S_t - goal||_2$. The reward function corresponds to the negative of the Euclidean distance between the particle and the goal location. The agent also receives a bonus reward of $+1$ when it is near the goal location.
\end{itemize}

Non-stationarity is introduced by either changing the location of walls or by randomly changing the latent goal location.

\subsection{Half-Cheetah in a Non-Stationary World}
This domain consists in a simulation of the high-dimensional \textit{Half-Cheetah} robot from OpenAI Gym (Brockman et al. 2016), using MuJoCo's physics engine (Todorov et al. 2012).
\textit{Half-Cheetah} agent is made up of 7 rigid links (1 for torso, 3 for forelimb, and 3 for hindlimb), connected by 6 joints.

\begin{itemize}
    \item Observation. $S_t \in \mathbb{R}^{17}$. The observed state is given by a 17-dimensional real-valued vector, including the position and velocity of the agent’s center of mass and the angular position and angular velocity of each of its 6 joints.
    \item Action. $A_t \in [-1.0,1.0]^6$. The action vector is the torque applied to each one of the agent's 6 joints.
    \item Reward. $R_t = -|v_{x} - v_{g}| -0.1 ||A_t||^2$, where $v_x$ is the agent's observed velocity along the x-axis, and $v_g$ is the target velocity. The agent's goal is to move forward while reaching a target-velocity and minimizing control costs.
\end{itemize}
We introduce three sources of non-stationarity:
    \begin{enumerate}
        \item \textbf{random wind}: an external latent horizontal force, opposite to the agent's movement direction, is applied;
        \item \textbf{joint malfunction}: either the torque applied to a joint of the robot has its polarity/sign changed; or a joint is randomly disabled;
        \item \textbf{target velocity}: the target velocity of the robot is sampled from the interval 1.5 to 2.5, causing a non-stationary change to the agent's reward function.
    \end{enumerate}

\section{Parameter Settings}

In Table~\ref{tbl:parameters} we show the parameters used for MBCD in the \textit{Half-Cheetah} and the \textit{Continuous Particle Maze} domains.

\begin{table}[h]
\caption{MBCD parameters.}
\label{tbl:parameters}
\centering
\begin{tabular}{|c|c|c|}
\hline
Environment                            & Half-Cheetah                       & Continuous Particle Maze          \\ \hline
$N$        & 5                                  & 5                                 \\ \hline
Dynamics architecture & 4 layers with 200 neurons          & 2 layers with 32 neurons          \\ \hline
$F$     & 250                                & 250                               \\ \hline
$L$     & $10^5$              & $10^5$            \\ \hline
Policy architecture      & 2 layers with 256 neurons & 2 layers with 64 neurons \\ \hline
$h$    & 100                                & 1000                              \\ \hline
$\delta$                    & 2                                  & 2.5                               \\ \hline
\end{tabular}
\end{table}

The neural networks used to model both the dynamics and the policies are Multi-Layer Perceptrons (MLP) with Rectified Linear Units (ReLU) activation function (Nair, 2010).
They were trained with mini-batch gradient descent using the Adam optimizer (Kingma \& Ba, 2015).

The parameters used for SAC and MBPO were the same (when applied) parameters as in Table~\ref{tbl:parameters}.

For GrBAL and ReBAL, we used a reference implementation provided by the authors\footnote{https://github.com/iclavera/learning\_to\_adapt}. 
The parameters used for the \textit{Half-Cheetah} domain were the same as in the original paper. 
A reduced number of layers and neurons were used for the \textit{Continuous Particle Maze} domain.
Regarding the action selection using MPC, we used the CEM method with 1000 candidate actions and planning horizon equal to 20 for all methods.

In all figures depicting our experimental results, each curve shows the mean and shaded areas present information about the standard deviation, when running the experiments with different random seeds.
We used $20$ different random seeds in all experiments, except for the experiments in Fig. 2 and Fig. 5 in which we used $7$ random seeds.

The code for MBCD is available at \url{https://github.com/LucasAlegre/mbcd}.